# HCDN: A Change Detection Network for Construction Housekeeping Using Feature Fusion and Large Vision Models

Kailai Sun[1], Zherui Shao[1], Yang Miang Goh*[1], Jing Tian[2], Vincent J.L. Gan[1]

*Corresponding author: bdggym@nus.edu.sg; Tel: +65 66012663

Contributing authors: skl23@nus.edu.sg;

[1]Department of the Built Environment, College of Design and Engineering, National University of Singapore, 4 Architecture Drive, Singapore 117566

[2]NUS-ISS, National University of Singapore, 25 Heng Mui Keng Terrace, Singapore 119615

# HCDN: A Change Detection Network for Construction Housekeeping Using Feature Fusion and Large Vision Models

**Abstract**

Workplace safety is critical as millions suffer from work-related accidents. Despite poor housekeeping contributing significantly to construction accidents, there's limited technological research focused on improving it. Recognizing and locating poor housekeeping in a dynamic construction site is important. Despite advances in AI and computer vision, existing methods for detecting poor housekeeping conditions face many challenges, including limited explanations, lack of locating of poor housekeeping and annotated datasets. To address these challenges, we propose the Housekeeping Change Detection Network (HCDN), an advanced change detection neural network that integrates a feature fusion module and a large vision model, achieving state-of-the-art performance. We introduce the approach to establish a novel change detection dataset (Housekeeping-CCD) focused on housekeeping in construction sites, along with a housekeeping segmentation dataset. Our contributions include significant performance improvements compared to existing methods, providing an effective tool for enhancing construction housekeeping and safety. We share our code and models: https://github.com/NUS-DBE/Housekeeping-CD.

**List of Abbreviations**

| | |
|---|---|
| AI | Artificial intelligence |
| AIGC | AI-Generated Content |
| CNN | Convolutional Neural Network |
| CV | Computer Vision |
| DL | Deep Learning |
| DNN | Deep Neural Network |
| ML | Machine Learning |
| DL | Deep Learning |
| NLP | Natural Language Processing |
| WSH | Workplace Safety and Health |
| CD | Change Detection |
| UAV | Unmanned Aerial Vehicle |
| ILO | International Labour Organization |
| CLIP | Contrastive Language-Image Pre-Training |
| CPU | Central Processing Unit |
| GPU | Graphics Processing Unit |
| MLP | Multilayer Perceptron |
| ViT | Vision Transformer |
| YOLO | You Only Look Once |
| SOTA | State Of The Art |
| IoU | Intersection over Union |
| SAM | Segment anything Model |
| R-CNN | Regions Convolutional Neural Network |
| SSD | Single Shot MultiBox Detector |
| MOTA | Multiple Object Tracking Accuracy |
| MOTP | Multiple Object Tracking Precision |
| YOLACT | You Only Look At Coefficients |
| CosLR | Cosine annealing Learning Rate |
| LVM | Large Vision Model |

# 1 Introduction

According to the International Labour Organization (ILO), every year, around 2.3 million workers globally lose their lives to work-related accidents and diseases [1], with about 360,000 of these being fatal incidents [2]. Additionally, nearly 337 million work-related accidents are recorded annually. The substantial operational disruptions caused by these injuries, along with the financial burden on societies, underscore the growing need for data-driven and artificial intelligence (AI) approaches to enhance safety management [3].

Housekeeping plays an important role in all workplaces, whether it is an office, factory, shipyard, construction site, storage facility, hospital, laboratory, retail shop, or industrial kitchen [4]. Workplace housekeeping may be defined as activities undertaken to create or maintain an orderly, clean, tidy, and safe working environment [5]. Workplace housekeeping is crucial for preventing injuries and accidents in the workplace [6]. Good housekeeping involves the efficient organization of the workplace, which significantly contributes to improved Workplace Safety and Health (WSH) performance, increased productivity, and better-quality control [7]. In contrast, poor housekeeping can result in workplace hazards that lead to many accidents, including slips, trips and falls, fire, struck by falling objects, and cutting by objects. Thus, ensuring good housekeeping in workplaces is critical to safety management.

Some governments are promoting the use of technologies to prevent injuries and accidents in high-risk workplaces. For example, the Singapore Manpower Ministry made it mandatory for construction worksites with contract sum more than S$5 million to install video surveillance system [8]. Another example is Smart Site Safety System (SSSS), including the use of video cameras and associated Artificial Intelligence methods, required by the Development Bureau in Hong Kong [9]. Thus, with these initiatives, video and image data become more prevalent, fueling increasing interest in the use of AI and computer vision (CV) technologies to improve WSH.

Despite the rapid development of AI and CV and the importance of housekeeping in construction sites, limited studies have tried to apply CV models to recognize poor housekeeping for accident preventions [10]. In addition, the causes and explanation of poor housekeeping using CV models have not been adequately developed due to several reasons. First, detecting poor housekeeping in the real world is complex, especially if the CV model is expected to explain the direct causes of poor housekeeping. Second, there is a lack of segmentation models to identify the presence or location of the objects (e.g., debris, water ponding) leading to poor housekeeping. Third, Machine Learning (ML) or Deep learning (DL) requires a lot of training data, but the publicly available construction site housekeeping datasets are not common. Fourth, there is a lack of high-accuracy Large Vision Model (LVM)-based CV models and network structures designed for housekeeping tasks. Thus, further research on the explanation and segmentation of construction housekeeping images is needed.

To address these issues, an advanced change detection neural network for housekeeping in construction sites is proposed, and a Housekeeping Construction Change Detection (Housekeeping-CCD) dataset creation method is proposed. In particular, a dataset filtering method is proposed to filter and clean the unmatched and weekly matched poor/good housekeeping pairs. Besides, we also establish a housekeeping segmentation dataset and compare state of the art (SOTA) methods, which can locate and segment the objects related to poor housekeeping. On the other hand, we designed a Housekeeping Change Detection

Network (named HCDN), which integrates proposed feature fusion module and Large Vision Model (LVM), achieving the SOTA performance on our dataset. The results underscore the effectiveness and potential for further applications.

This study aims to mitigate construction housekeeping problems by providing a novel change detection approach to detect and explain poor housekeeping images. The main contributions of this study are as follows: (1) We propose a change detection dataset creation pipeline for facilitating research on CV solutions to mitigate housekeeping problems in construction sites. (2) A dataset filtering method to automatically filter and clean the raw images captured from real-world constructions is introduced. (3) We designed a novel Housekeeping Change Detection Network (HCDN) with Large Vision Model (LVM) and more effective features fusion, achieving superior performance than existing SOTA methods. (4) We conducted an ablation study to demonstrate that our Housekeeping-CCD dataset can effectively be used as an image segmentation dataset for identifying causes of poor housekeeping, since the model performs well even when only the poor housekeeping image is provided. (5) We integrate HCDN into the proposed housekeeping detection system, which shows its potential in practical construction site situations. The results underscore the effectiveness and potential of our housekeeping change detection dataset and methods for real world applications. To promote further development, we share our source code and trained models for global researchers: https://github.com/NUS-DBE/Housekeeping-CD.We believe this study provides the foundation for facilitating further research in construction CV applications.

The remainder of the paper is organized as follows. Section 2 first describes the related works, followed by the introduction of change detection algorithms commonly used in the building field. In Section 3, we introduce how to collect and annotate the real-world housekeeping change detection datasets. In Section 4, we introduce the proposed housekeeping change detection network. The experimental procedure followed to test performance in detecting construction change is described in Section 4. Afterwards, Section 5 presents the results of the proposed network trained on our dataset, compared with SOTA change detection methods. Consequently, Section 6 discusses the performance of change detectors and presents the practical applications in real-world construction sites. Finally, conclusions and future work are provided in Section 7.

## 2 Related work

### 2.1 Generic object detection datasets in construction

Using CV to automatically monitor construction sites can enhance safety, quality, and productivity. These CV systems rely on the ability to detect construction workers, equipment, and other objects in images and videos. Hence, some studies aim to detect generic objects in construction (e.g., workers, vest, helmet, wood, and scaffold) and a few open-source construction datasets have been published. Moving Objects in Construction Sites (MOCS) dataset [11] has 41,668 images collected from 174 construction sites. MOCS dataset is annotated with 13 categories of moving objects and 222,861 instances, including tower crane, hanging hook, vehicle crane, roller, bulldozer, excavator, loader, pump truck, concrete transport mixer, pile driver, etc. They trained and compared 15 different DNN-based detectors (e.g., YOLO, Faster RCNN, SSD) on the MOCS dataset, achieving about 40% mAP. Site Object Detection Dataset (SODA) [12] has 19,846 images, including 286,201 objects with 15 categories, including board, handcart, fence, wood, cutter, slogan, rebar, ebox, brick, hopper, hook, etc. They used an UAV to capture different equipment at different angles and times on many construction sites in China. They trained and compared YOLOv3 and YOLOv4 on their SODA to achieve high performance. The Construction Instance Segmentation (CIS) [13] contains 50k images, including more than 83k annotated instances with 10 categories. CIS dataset includes 2 categories of workers (workers wearing & not wearing safety helmets); 1 categories of materials (precast components);7 categories of machines (precast components delivery trucks, dump trucks, concrete mixer trucks, excavators, rollers, dozers & wheel loaders). Wang et al. [14] developed an object detection dataset, which includes 7,500 images with 9 categories, including dust, spraying car, mulching fabric, spraying sink, etc. They improved YOLOv7 with deformable convolutional networks and wise-IoU to improve YOLOv7 to achieve high precision (mAP = 68.1%). Alberta Construction Image Dataset (ACID) [15] is developed specially to identify construction machinery. They manually collected and annotated 10,000 images of 10 categories of construction machines, including excavator, compactor, dozer, grader, dump truck, concrete mixer truck, wheel loader, backhoe loader, tower crane, and mobile crane. They trained and compared YOLOv3, Faster RCNN, SSD methods on their ACID dataset, achieving about 89.2% mAP performance. In summary, even though global researchers have made efforts to improve generic object detection in construction, there is a lack of housekeeping dataset, which makes automatic detection of housekeeping issues difficult.

## 2.2 Generic object segmentation in construction

Many existing studies use segmentation methods in construction sites. Xiao et al. [16] applied Mask R-CNN to segment workers and then used Kalman filters to track workers in construction sites. Compared with existing methods, their method achieves the SOTA tracking performance with multiple objects tracking accuracy (MOTA) of 96.4%, multiple object tracking precision (MOTP) of 86.2%, and F1-score of 97.6%. In vision-based monitoring outdoor works in construction sites, there is a challenge in developing robust object detection and segmentation methods that are reliable in different weather conditions. To address this challenge, Kang et al. [17] trained the You Only Look At Coefficients (YOLACT) model and propose weather augmentation algorithms to deal with the five weather conditions: brightness, darkness, rain, snow, and fog, achieving about 5% mAP improvement. To improve building waste management, another study [18] collected a dataset of construction waste (5366 images) and trains the DeepLabv3+ model for accurate waste segmentation, achieving a mean Intersection over Union (mIoU) of 0.56. They considered that the waste includes different types: debris, plastic, rock, stone, rubble, gravel, wood, etc. Another study [19] compared the different segmentation methods (e.g., U-net and DeepLabv3+) with different backbone networks (ResNet-101, RegNetX-1.6, MobileOne-S3) for recognizing construction, renovation, and demolition solid waste in-the-wild, achieving the good performance (0.85 mean pixel accuracy and 0.74 frequency-weighted IoU). Hwang et al [20] applied a Mobilenet-based image segmentation model for automatically annotating the foreground and the background of construction images. Woldeamanuel et al. [21] trained a deep learning-based image segmentation model with thermal and RGB images for estimation of concrete strength and safety management at construction sites, achieving about 80% prediction accuracy.

A study [22] applied vision-based 3D/2D segmentation methods and compares different backbone networks to segment the wind-borne debris for disaster preparedness. Ji et al. [23] proposed an encoder-decoder 3D point cloud segmentation model with a feature fusion algorithm to discover seepage for further maintaining the healthy conditions of the tunnel, achieving superior performance than other segmentation algorithms. Pal et al. [24] trained a deep learning-based semantic segmentation method (Mask R-CNN) with their custom construction dataset for construction activity-level (e.g., formwork, reinforcement, concrete placement) progress monitoring, achieving about 0.9 mAP (IoU≥0.5) and 0.78 mAP (IoU≥0.75). In summary, even though global researchers have made efforts to develop object segmentation datasets and methods in construction, there is a lack of object segmentation for the specific housekeeping problem.

## 2.3 Change detection in construction

Change detection is important for monitoring building changes and deformations, construction progress monitoring, disaster management, etc. For instance, many existing studies [25], [26], [27] apply 3D change detection methods in construction progress monitoring. Current progress monitoring mostly relies on visual inspections performed by workers, which are time and cost-intensive surveys. In contrast, 3D change detection method is a good alternative for accurate and fast "as-built" recording on site. Meyer et al. [26] propose an accurate change detection method based on dense 3D point clouds and a new metric to evaluate the accuracy of a BIM. Huang et al, [25] analyze the relationship between point-based changes, voxel- or occupancy grid-based changes, and segment/object-based changes. Another study [28] applies an auto encoder network to complete the point clouds for building change detection. Another application is city-scale development tracking [29]. They capture bi-temporal LIDAR data and develop change detection methods for differentiating between new and demolished buildings.

Besides, change detection methods are well developed in remote sensing [30]. To detect the changed buildings and "where" the change has occurred, many studies use drones and satellite to capture the earth images at different times. There are many change detection datasets, e.g., LEVIR-CD [31], WHU-CD [32], S2Looking [33]. LEVIR-CD dataset has 637 high-resolution Google Earth image patch pairs with a size of 1024 × 1024 pixels. WHU-CD dataset has 8,189 tiles with 512×512 pixels. S2Looking dataset has 5,000 registered bi-temporal image pairs with 1024 × 1024 pixels of rural areas and more than 65,920 annotated change instances. Indeed, there are many change detection methods. Changer [34] uses Aggregation-Distribution (AD) and features "exchange" modules for bi-temporal image analysis. LightCDnet [35] is a compact change detection model that efficiently preserves input data through an early fusion backbone and pyramid decoder, achieving good performance while being significantly smaller in size. spatial-temporal attention neural network (STANet) [31] introduces a Siamese-based network with a CD self-attention mechanism to analyze spatial-temporal dynamics and includes the LEVIR-CD dataset with 637 image pairs. Time Travelling Pixels (TTP) [36] integrates SAM foundation model knowledge into change detection. Bi-Temporal Adapter Network (BAN) [37] incorporates a CV foundation model, Bi-TAB, and bridging modules, achieving top performance on multiple datasets such as LEVIR-CD and Landsat-SCD. In summary, even though global researchers have made efforts in building change detection, there is a lack of change detection systems for housekeeping.

## 3 Dataset

As previously mentioned, this paper introduces a deep neural network and change detection dataset developed specifically for construction housekeeping. This section presents our dataset, named Housekeeping Construction Change Detection (Housekeeping-CCD), a novel resource for change detection tasks. Housekeeping-CCD focuses on the housekeeping problem in construction sites. Unlike remote sensing datasets, RGB images in Housekeeping-CCD are captured by body-worn cameras or cell phones, which are common and widely used in practice. Images in Housekeeping-CCD are captured with diverse viewpoints in complex scenes. We use consumer-grade body-worn cameras and phones, with shooting heights ranging from 0.5m to 3m and diverse shooting angles. Unlike 3D cameras, body-worn cameras are more widely used by workers in construction sites. Our dataset is sourced from actual construction sites, using images taken by construction site personnel. Therefore, it can be quickly and easily applied on construction sites without additional camera deployment and data collection work. The dataset creation details are introduced in the following subsections.

### 3.1 Data acquisition

The raw image data was provided by the Singapore Housing & Development Board (HDB). The dataset contains 1,864 pairs of images. Each pair includes a photo of a poor housekeeping scene and a corresponding photo taken at the same location after the issue has been resolved. These image pairs were collected from 40 construction sites across Singapore between March 2023 and June 2023. All images were captured by site personnel using either body-worn cameras or cell phones.

### 3.2 Data annotation

To develop our high-quality dataset, we initially employed over three annotators from a professional annotation company. They spent approximately two weeks labeling the data to create segmentation masks. Because the poor housekeeping is very complex, generic object detection bounding box is not suitable for this task. Thus, we use segmentation mask, instead. These masks are crucial as they define the boundaries of complex and diverse objects within the construction.

Following the initial annotation phase, we engaged three experts—comprising two researchers with significant experience in data annotation and one professor in this field—to review the annotated dataset. This process involved two iterative rounds where the dataset annotations were carefully checked and the incorrect annotations returned to the original annotators for corrections. The cycle (annotation, expert review, and correction) ensured that our dataset achieved a high level of accuracy and reliability.

## 3.3 Data filtering method

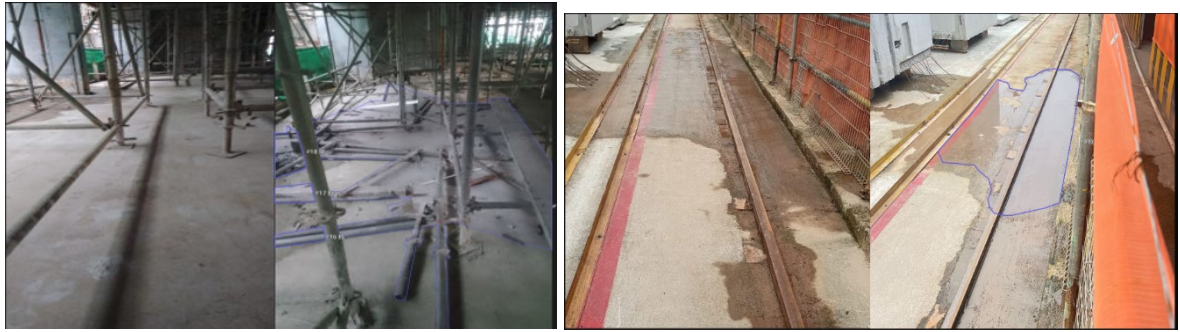

(a) Different camera distances  (b) Different horizontal views

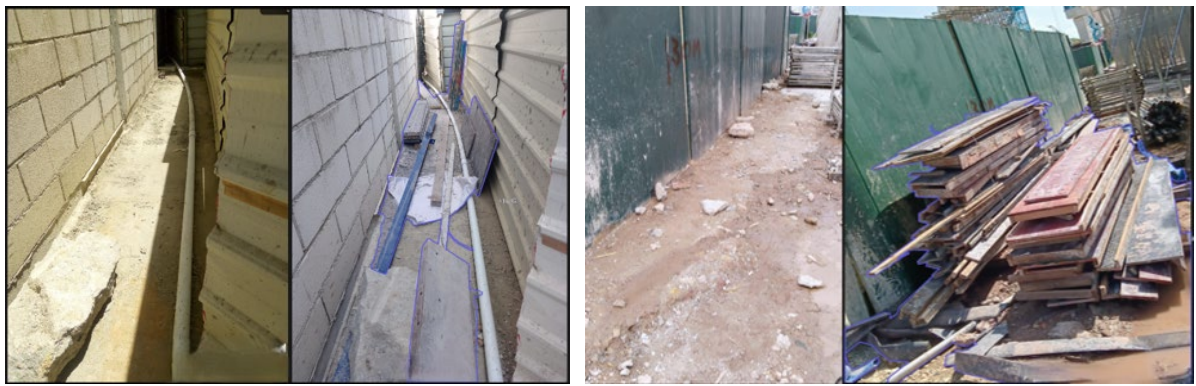

(c) Different lighting conditions  (d) Different rotational views

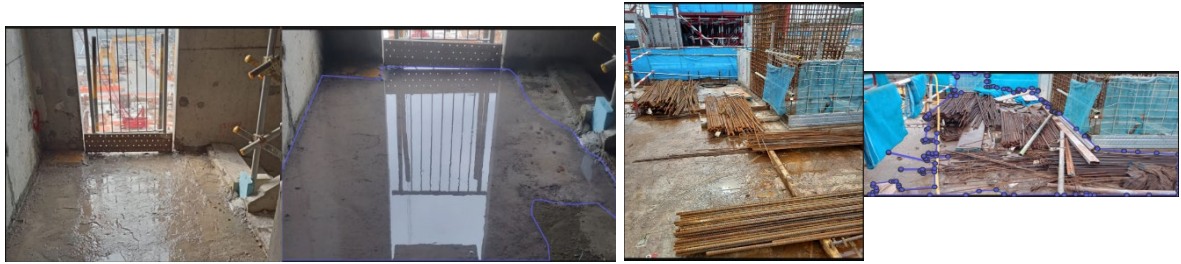

(e) Different vertical views  (f) Different image sizes

Fig.1. Visualization of some challenges in the raw dataset. For each group, workers take two images (poor housekeeping image and good housekeeping image) in construction sites.

As part of housekeeping monitoring processes, site personnel take images of the poor housekeeping and then an image of the good housekeeping after workers have rectified the poor housekeeping. These pairs of images can be used to train change detection models. However, the pairs of images are taken in an open environment and there are many challenges: different lighting conditions, horizontal views, vertical views, rotational views, image sizes, and camera distances (Fig.1). These challenges cause difficulty in detecting the object changes across each pair of housekeeping images. Thus, we propose a data filtering method to obtain high-quality image pairs in Fig.2. First, we use SIFT (Scale-Invariant Feature Transform) to extract features across a pair of good housekeeping and poor housekeeping images. SIFT is used to detect and describe local features in images, which are invariant to scale and rotation. Second, the extracted SIFT features are then compared across images to find matching pairs of points. The flann matcher with k-nearest neighbor method is employed to match the feature points from good and poor housekeeping images. Third, after feature matching, we decided the matched pairs and mismatched pairs based on the distance threshold of 0.7, i.e., if the matching score is greater than 0.7, we consider it a matched pair; otherwise, it is classified as a mismatched pair. The mismatched pairs are discarded to ensure only correct feature correspondences are used in the next steps. Fourth, the good housekeeping image is aligned based on the matched pairs of features. We transform the good housekeeping image with a rotation matrix and translation matrix so that the matched features are aligned as closely as possible. In this step, only the good image is transformed, while the poor housekeeping image remains unchanged. Fifth, we should not accept the big angle differences between good and poor housekeeping images. Thus, after alignment, we check if the alignment leaves a blank area in the images that is less than 20% of the total image area. Last, we select the "Good Match" image pairs, which will be fed into the change detection network.

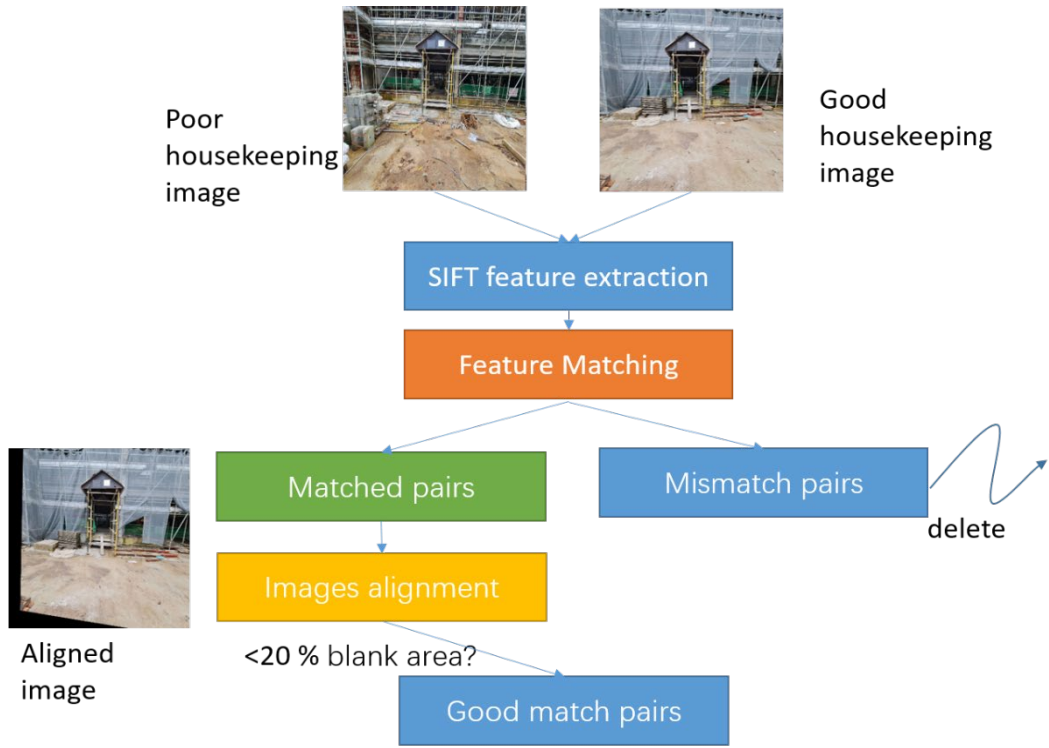

Fig.2 The overall method of data filtering

The raw dataset has 1864 image pairs. After data filtering and alignment, we obtained 700 images. We named the filtered and aligned dataset Housekeeping-CCD. We visualize some examples in Fig.3. In each sub image, the left top shows the raw poor housekeeping image; the right top shows the raw good housekeeping image; the left bottom shows the ground truth; the right bottom shows the aligned good housekeeping image. As we can see, the cause of poor housekeeping (ground truth) is diverse and complex. The housekeeping change detection is an important task, which takes bi-temporal images as input and predicts "where" the change has occurred.

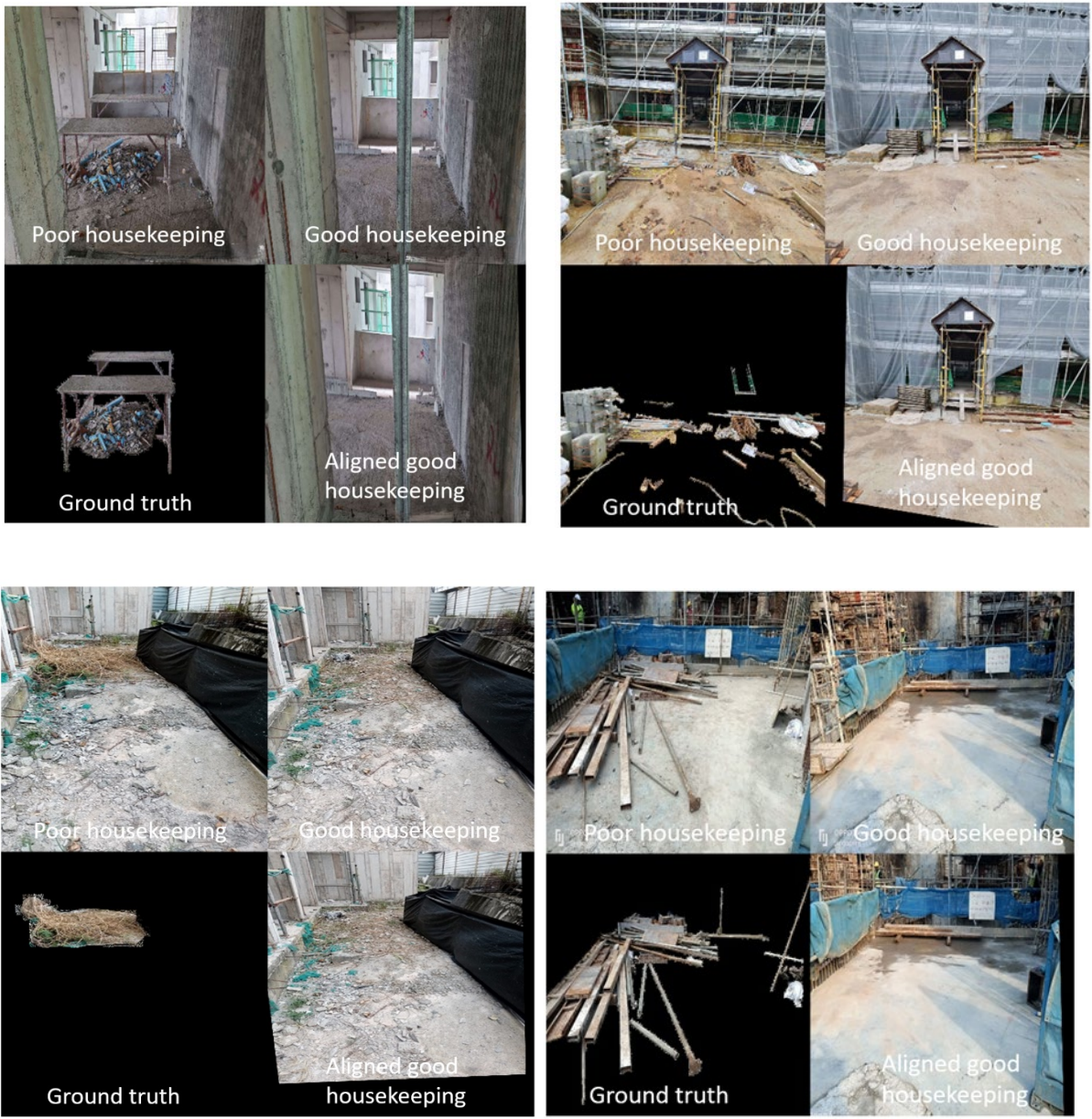

Fig.3 Dataset sample visualization.

### 3.4 Dataset Statistical indicators

After data annotation and filtering, our dataset has 186 indoor image pairs, and 514 outdoor image pairs. To further analyze our dataset, we count the poor housekeeping type in Fig.4. Our dataset only focuses on five causes (Debris, rebars, steel pipes, water ponding and others) for poor housekeeping. It highlights that "Water Ponding" and "Debris" are the most common types of poor housekeeping issues (233, 182), while "Others" is the least common (42).

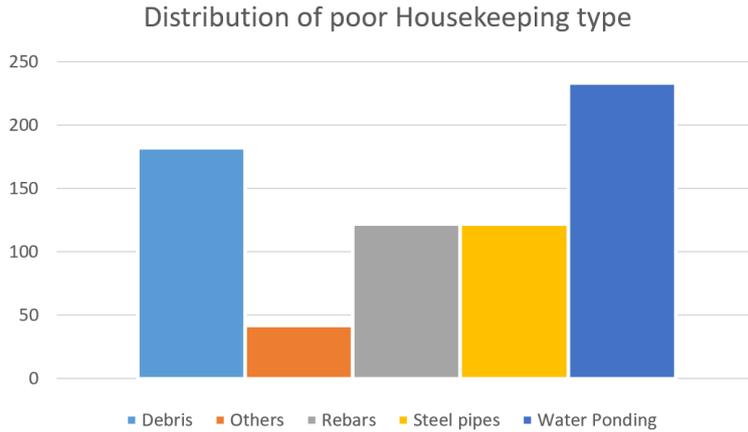

Fig 4. Distribution of poor housekeeping type

Besides, we visualize the distribution (percentage) of change area at the pixel level in Fig.5. To achieve this, first, we compute the change area in terms of number of pixels for each image pair. Specifically, the change area refers to the total number of pixels that have changed between two images in a pair. The change area is also the ground truth area in Fig.3. Because the change area differs between image pairs, we normalize the change area as change area ratio. Change area ratio can be defined as a relative measure that expresses the change area as a percentage of the total area of the image, providing a normalized comparisons across image pairs. For example, a 20% change area ratio means that the change area constitutes about 20% of the total area of the image. The histogram in Fig. 5 (a) illustrates the distribution of these change area ratios in our dataset. In Fig.5 (a), we find that most of the images fall within the 20%-30% and 30%-40% change area ratio ranges. This suggests that 20%-40% changes are the most common in our dataset. As the change area ratio increases beyond 40%, the percentage gradually decreases. The occurrence of changes with ratios above 70% is quite rare, with the percentages dropping to below 5%. It often occurs with the "water ponding" poor housekeep type in Fig.5 (c). Also, we find the difference between distributions based on different poor housekeeping types. In Fig.5 (b), the highest percentage falls within the 10%-20% change area ratio ranges with the "Debris" poor housekeeping types; in Fig.5 (c), the highest percentage falls within the 20%-40% change area ratio ranges with the "water ponding" poor housekeeping types; in Fig.5 (d), the highest percentage falls within the 10%-20% change area ratio ranges with the "Steel pipes" poor housekeeping types; in Fig.5 (e), the highest percentage falls within the 10%-30% change area ratio ranges with the "Rebars" poor housekeeping types; in Fig.5 (f), the highest percentage falls within the 10% change area ratio ranges with the "Others" poor housekeeping types.

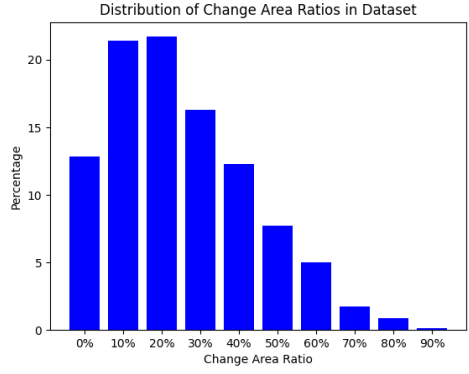
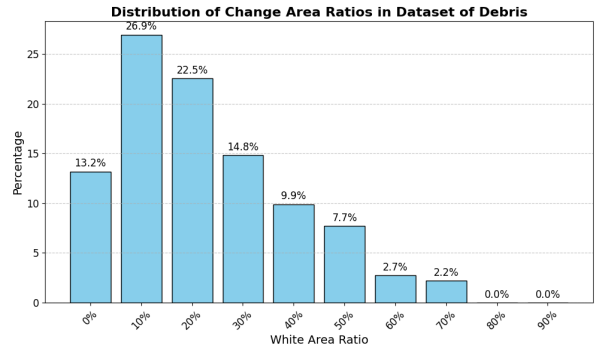

(a)                  (b)

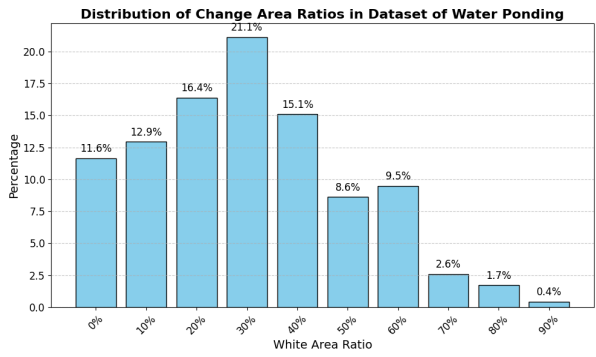
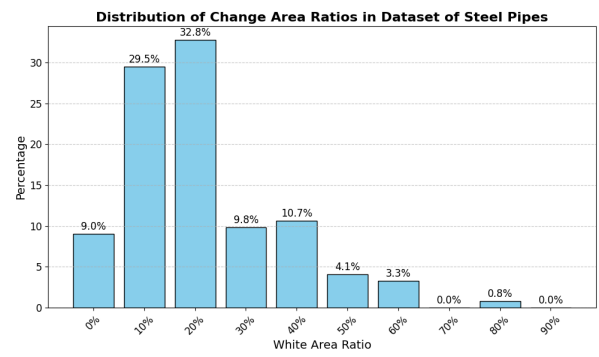

(c)                  (d)

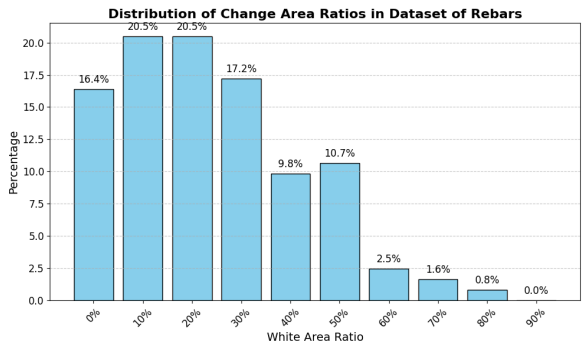
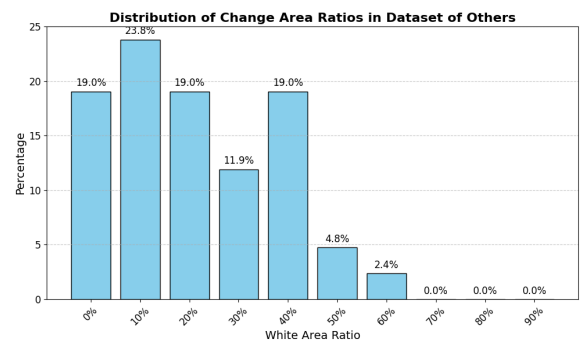

(e)                  (f)

Fig 5. The distribution of change area ratios.

## 3.5 Dataset split

We split Housekeeping-CCD into three parts with 7:2:1 ratio, i.e., the training set includes 490 image pairs (good housekeeping image, poor housekeeping image, label), validation set includes 140 image pairs (good housekeeping image, poor housekeeping image, label), and the test set includes 70 image pairs (good housekeeping image, poor housekeeping image, label).

## 4 Housekeeping change detection network (HCDN)

After preparing the dataset, a Housekeeping Change Detection Network (HCDN) is developed. HCDN encompasses a change detection model that integrates a feature fusion module and a large visual model to achieve superior performance. The structure is discussed in the following subsections.

### 4.1 The overall structure of HCDN

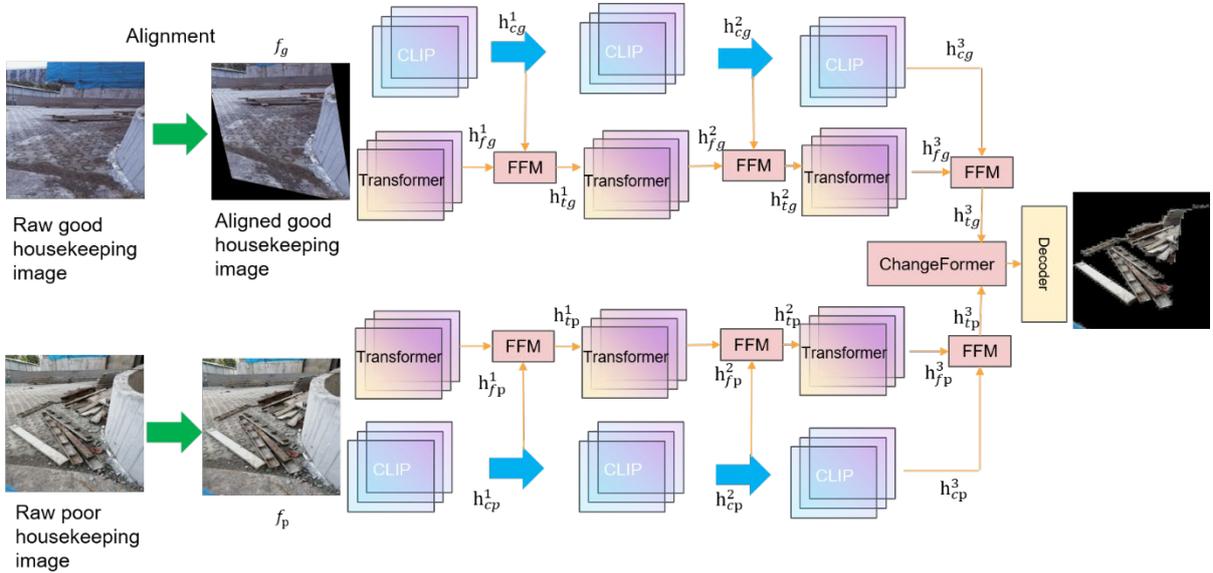

Fig 6. The overall structure of HCDN. "CLIP" means the image encoder of CLIP.

Fig 6 shows the overall structure of HCDN. Given the raw poor housekeeping image $f_p$ and the aligned good housekeeping image $f_g$, we feed them into our HCDN. Our HCDN has a Siamese network structure with two identical subnetworks that share the same weights. $f_g$, $f_p$ are processed through these identical networks to determine their similarity or difference. Firstly, we employ the ViT backbone of a visual foundational model, named Contrastive Language–Image Pre-training (CLIP) to extract the image latent features $h_c = \{h_{cp}, h_{cg}\}$:

$$h_{cp}, h_{cg} = \text{ViT}_{\text{CLIP}}(f_p), \text{ViT}_{\text{CLIP}}(f_g). \qquad (1)$$

CLIP [38] is a multimodal model developed by OpenAI. It uses contrastive learning to align images with their corresponding text, enabling zero-shot learning. CLIP can perform tasks like image search and content moderation across diverse domains. Then, we employ the Transformer block [39] to extract another image feature $h_f = \{h_{fp}, h_{fg}\}$:

$$h_{fp}, h_{fg} = \text{Transformer}(f_p), \text{Transformer}(f_g). \tag{2}$$

Then, we design a Feature Fusion Module (FFM) to fuse two features:

$$h_{tp}, h_{tg} = \text{FFM}(h_{cp}, h_{fp}), \text{FFM}(h_{cg}, h_{fg}). \tag{3}$$

The above process will be repeated three times to gradually extract and fuse the effective features. Next, we employ Changformer [40] and a decoder to decode the feature into a predicted mask image $\hat{y}$:

$$\hat{y}(m, n) = \text{Decoder}\big(\text{Changformer}(h_{tp}, h_{tg})\big), \tag{4}$$

where $\hat{y}(m, n)$ denotes the predicted output value at position (m,n) in an image. The output value of predicted mask image $\hat{y}$ presents the classes i.e., change or no-change.

**4.2 Feature Fusion Module (FFM)**

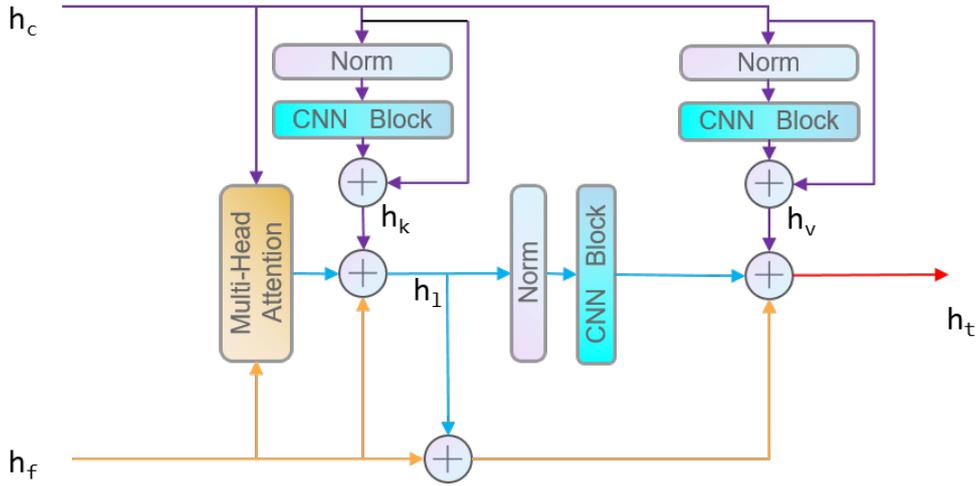

Fig 7. The detailed structure of Feature Fusion Module (FFM)

In this section, we will introduce the proposed Feature Fusion Module (FFM). Inspired by [41], the feature $h_c$ from CLIP and the feature $h_f$ from Transformer will be fused effectively to improve the accuracy of our proposed method. As Fig 7 shows, the FFM includes a multi-head attention:

$$\text{Attention}(\boldsymbol{Q}, \boldsymbol{K}, \boldsymbol{V}) = \text{Softmax}\left(\frac{\boldsymbol{QK}^T}{\sqrt{d_{\text{head}}}}\right)\boldsymbol{V}, \tag{5}$$

where Q, K and V denote Query, Key and Value, respectively. Here, we consider $h_f$ as Q, and $h_c$ as K and V. We can thus obtain:

$$\text{Attention}(h_f, h_c, h_c) = \text{Softmax}\left(\frac{h_f h_c^T}{\sqrt{d_{\text{head}}}}\right)h_c, \tag{6}$$

Then, we compute the $h_k$:

$$h_k = \text{CNN}(\text{Norm}(h_c)) + h_c, \tag{7}$$

where we use layernorm function as Norm function, and we use a simple 3-layer CNN as our CNN block. Next, we obtain the $h_l$:

$$h_l = h_f + h_k + \text{Attention}(h_f, h_c, h_c). \tag{8}$$

It means that we not only consider the original features $h_f$, $h_c$, but also consider the relation between them ($Attention(h_f, h_c, h_c)$). Lastly, we conduct the second feature fusion:

$$h_t = h_l + h_v + h_f + \text{CNN}(\text{Norm}(h_l)). \tag{9}$$

It means that we not only consider the original features $h_f$, $h_c$, but also consider the relation between them $h_l$. After feature fusion, the information of poor housekeeping image and good housekeeping image are extracted deeply. The proposed feature fusion module learns the optimal distance metric at each scale during training- resulting in better change detection performance.

### 4.3 Loss function

The HCDN employs a weighted cross entropy function as follows:

$$L = -\sum_{i=0}^{N} w * y_i(m, n) \log\left(\hat{y}_i(m, n)\right) + (1 - w)(1 - y_i(m, n)) \log(1 - \hat{y}_i(m, n)), \tag{10}$$

where N is the number of training samples in the dataset. $y_i(m, n)$ is the ground truth label (0 or 1) for pixel (m,n). $w \in [0,1]$ is the weight that is assigned to the positive class, helping to address class imbalance by giving more importance to the minority class.

# 5 Experiment

To test the image dataset and proposed HCDN, an experiment was conducted. Subsections below discuss the evaluation indicator, experiment environment, and results and discussion.

## 5.1 Evaluation Indicators

Standard metrics for image segmentation tasks were employed in this study. First, precision measures the proportion of predicted positive pixels (i.e., pixels classified as belonging to the target region in the ground-truth image) that are actually positive:

$$\text{Precision} = \frac{\text{TP}}{\text{TP} + \text{FP}}, \tag{11}$$

where TP (True Positive) is the number of pixels correctly predicted as the target class, and FP (False Positive) is the number of pixels incorrectly predicted as the target class.

Then, recall measures the proportion of actual positive pixels that are correctly predicted by our model.

$$\text{Recall} = \frac{\text{TP}}{\text{TP} + \text{FN}}. \tag{12}$$

FN (False Negative) is the number of pixels that were incorrectly predicted as the background class when they should have been classified as the target class.

Next, F1 score is the harmonic mean of precision and recall. It provides a balance between precision and recall:

$$\text{F1 score} = 2 \times \frac{\text{Precision} \times \text{Recall}}{\text{Precision} + \text{Recall}}. \tag{13}$$

Next, IoU (Intersection over Union) measures the overlap between the predicted segmentation and the ground truth segmentation:

$$\text{IoU} = \frac{\text{Intersection Area}}{\text{Union Area}} = \frac{\text{TP}}{\text{TP} + \text{FP} + \text{FN}}. \tag{14}$$

Intersection area is the area where the predicted region and the actual region overlap (i.e., the true positive area, TP). Union area is the area covered by either the predicted region or the actual region, or both (including TP, FP, and FN).

Last, accuracy (Acc) measures the proportion of all pixels that are correctly classified:

$$\text{Acc} = \frac{TP + TN}{TP + TN + FP + FN}, \quad (15)$$

where TN (True Negative) is the number of pixels correctly predicted as the background class.

Because this change detection task is a two-class classification task, we also employ the two-class average/mean metrics (i.e., aACC, mF-score, mPrecision, mRecall, mIoU) to measure CD performance.

## 5.2 Experimental environment

Experiments are conducted on a computer with 64- bit Linux 22.04 platform, Intel(R) Xeon(R) Gold 6438N 128-Core Processor, 500 GB RAM, and 8 NVIDIA H100 GPUs (80G display memory) with CUDA version 12.0. The programming platform is Python 3.10, and the deep learning library is Pytorch 1.13.

## 5.3 Hyper-parameters

The input images are resized to 1024x1024.The training process is accelerated by using the Adam optimizer. We set the Weight $w$ as 0.3, which is consistent in section 3.4. We adopt the cosine annealing learning rate (CosLR) scheduler with an initial learning rate of 3e-4. The batchsize is set to 8. We train our model for a total step of 100K.

In our experiment, we applied many data preprocessing and augmentation techniques to improve data diversity. The following augmentation strategies were implemented during training. (1) Images were randomly rotated within a range of ±180 degrees with a probability of 50%. (2) We performed random cropping on the images to a fixed size of 256×256 pixels, ensuring that no single class occupied more than 75% of the cropped area. (3) We applied data normalization to the input images using specific mean and standard deviation values to standardize the pixel intensity distributions across the dataset. The mean values were [122.8,116.7,104.1] and the standard deviation values were [68.5,66.6,70.3]. (4) To further increase data diversity, images were horizontally and vertically flipped independently with a probability of 50%. (5) We applied random photometric adjustments to the images, including modifications in brightness (±10 units), contrast (ranging from 0.8 to 1.2 times the original), saturation (ranging from 0.8 to 1.2 times the original), and hue (±10 degrees). These augmentation techniques can help the model generalize better by exposing it to different spatial configurations and various lighting conditions.

## 5.4 Main results

We compare our methods with baseline SOTA methods. Changer [34] introduces aggregation-distribution (AD) and feature "exchange" modules to learn the bi-temporal features between two images. LightCDnet [35] is a lightweight change detection model designed to efficiently preserve input information through its early fusion backbone network and pyramid decoder. It achieves good performance while being 10 to 117 times smaller in model size. STANet [31] is a novel Siamese-based spatial–temporal attention neural network with a CD self-attention mechanism to model the spatial–temporal relationships. The authors also introduce a famous CD dataset (LEVIR-CD) with 637 image pairs (1024 × 1024). Time Travelling Pixels (TTP) [36] integrates the latent knowledge of the SAM foundation model into change detection task. BAN [37] integrates a CV foundation model, bi-temporal adapter branch (Bi-TAB), and bridging modules into change detection task, achieving the SOTA performance on many change detection datasets (e.g., LEVIR-CD dataset, BANDON dataset, The Landsat-SCD dataset).

Tab.2 Comparison against SOTA CD methods on the validation set of Housekeeping-CCD

| Metric (Validation set) | Year | Backbone | Lr schd | aACC | mF-score | mPrecision | mRecall | mIoU |
|---|---|---|---|---|---|---|---|---|
| Changer [34] | 2023 | Resnet-18 | 40000 | 82.19 | 80.12 | 79.18 | 82.4 | 67.42 |
| Changeformer [40] | 2022 | MIT-B1 | 40000 | 85.91 | 83.26 | 83.52 | 83.02 | 71.88 |
| STANet [31] | 2020 | BAM | 40000 | 79.07 | 72.72 | 76.7 | 71.05 | 58.83 |
| LightCDnet [35] | 2023 | Custom-large | 40000 | 73.75 | 72.86 | 74.3 | 78.63 | 57.54 |
| BAN [37] | 2024 | ViT-L, Clip | 40000 | 86.69 | 84.93 | 83.89 | 86.53 | 74.16 |
| TTP [36] | 2024 | ViT-SAM-l | 300 epoch | 85.89 | 84.41 | 83.18 | 84.93 | 73.32 |
| Ours | 2024 | Clip, FFM | 40000 | **88.25** | **86.30** | **85.91** | **86.74** | **76.27** |

Tab.3 Comparison against SOTA CD methods on the test set of Housekeeping-CCD

| Metric (test set) | year | Backbone | Lr schd | aACC | mF-score | mPrecision | mRecall | mIoU |
|---|---|---|---|---|---|---|---|---|
| Changer | 2023 | Resnet-18 | 40000 | 83.25 | 80.24 | 78.78 | 83.02 | 67.69 |
| Changeformer | 2022 | MIT-B1 | 40000 | 86.62 | 83.22 | 82.72 | 83.77 | 71.98 |
| STANet | 2020 | BAM | 40000 | 81.31 | 74.00 | 77.08 | 72.28 | 60.65 |
| LightCDnet | 2023 | Custom-large | 40000 | 72.73 | 71.19 | 72.8 | 79.08 | 55.69 |
| BAN | 2024 | ViT-L, Clip | 40000 | 87.99 | 85.55 | 84.04 | **87.88** | 75.23 |
| TTP | 2024 | ViT-SAM-l | 300e | 85.54 | 82.87 | 81.29 | 85.69 | 71.32 |
| Ours | 2024 | Clip, FFM | 40000 | **89.32** | **86.67** | **85.95** | 87.50 | **76.97** |

In Tab.2, we train and evaluate our proposed model, HCDN, alongside other state-of-the-art (SOTA) methods on our validation dataset. The results show that HCDN outperforms all prior methods across every metric, making it the most effective model in this comparison. The BAN model with a ViT-L backbone emerges as the second-best performer, demonstrating consistently strong results but falling slightly short of HCDN. Notably, HCDN achieves significant improvements in Accuracy (2.84%) and Recall (2.41%) compared to BAN.

In Tab.3, we evaluate our proposed model, HCDN, against other SOTA methods on the test dataset. The results indicate that HCDN remains the top-performing model across all key metrics, confirming its effectiveness in comparison to prior methods. BAN with the ViT-L backbone secures the second position, demonstrating strong performance, especially on the mRecall metric. Besides, HCDN achieves significant improvements in overall Accuracy (1.51%) and mIoU (1.74%) compared to BAN, solidifying its superiority in change detection tasks.

We visualize some results in Fig.8. As we can see, the poor housekeeping causes (ground truth) is diverse and complex. Overall, our HCDN can predict the results accurately. These underscore observations that the method proposed in this paper can effectively bridge the domain gap and enhance housekeeping change detection understanding. But HCDN still makes some small errors, especially for water ponding images, which leaves room for future research.

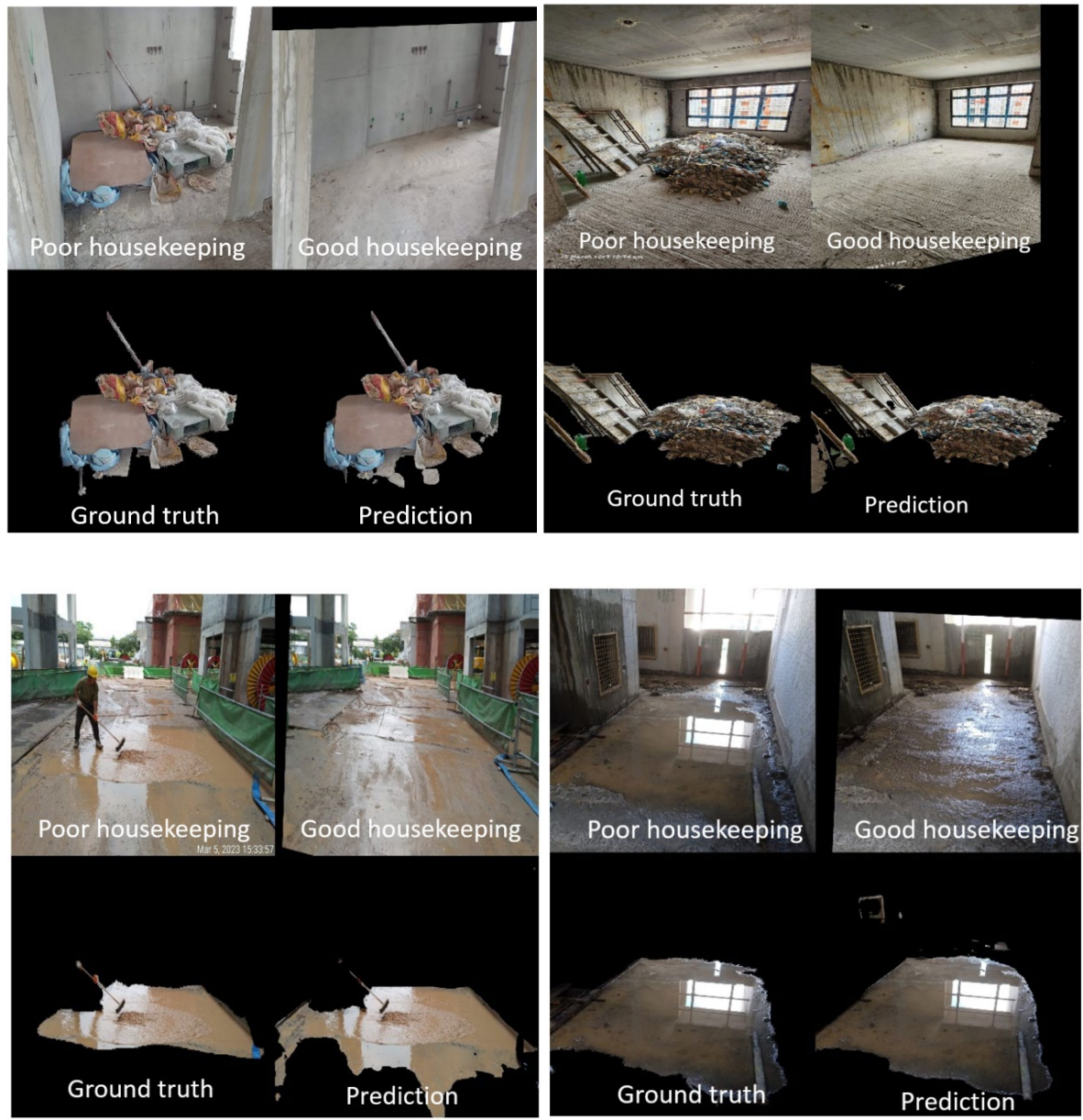

Fig 8. Qualitative results of HCDN on test set. In each sub image, the left top shows the raw poor housekeeping image; the right top shows the raw good housekeeping image; the left bottom shows the ground truth; the right bottom shows the predicted results by HCDN.

## 5.5 Ablation experiment

To thoroughly evaluate the effectiveness and importance of good housekeeping image, we conducted a series ablation experiment to explore the difference. To achieve it, we mask good housekeeping image as a black image in Fig.9. We feed the poor housekeeping image and this whole black image into our HCDN. Then, we retrain our HCND with the same network and hyper-parameters with the masked image. Last, we compare the results in Tab.4. The performance experienced some declines when we masked good housekeeping images. The overall average accuracy drops only slightly by 0.34%, but the mF1-score, mPrecision, mRecall, and mIoU decrease by 1.96%, 1.0%, 2.91%, and 2.62%, respectively. The declines mean that the good housekeeping images are important in this process.

In summary, the ablation experiment confirms that the absence of good housekeeping image will lead to a significant decline in performance, especially in terms of mRecall and mIoU, which are critical for identifying true positives and accurately segmenting the change area. It also validates the importance of good housekeeping image in the housekeeping change detection task.

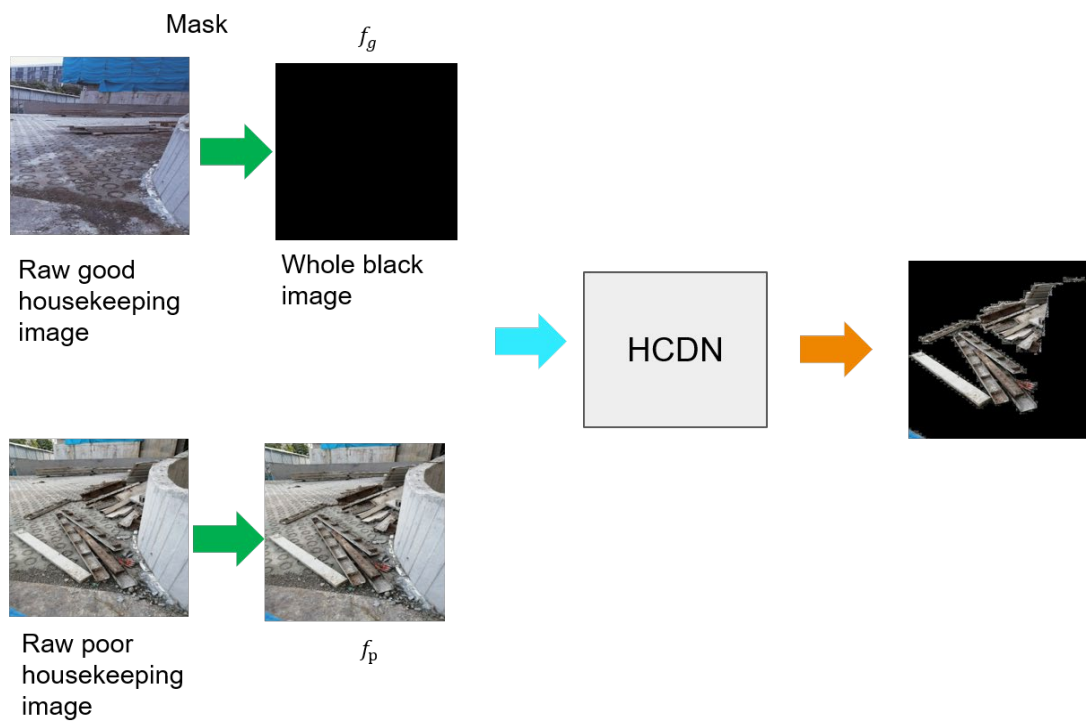

Fig 9. The overall method flowchart in the ablation experiment.

Table. 4 The CD performance with different inputs

| Metric | aACC | mF-score | mPrecision | mRecall | mIoU |
|---|---|---|---|---|---|
| Input with All images | 88.25 | 86.30 | 85.91 | 86.74 | 76.27 |
| Input without good housekeeping images | 87.91 | 84.34 | 84.91 | 83.83 | 73.65 |
| Difference | **0.34** | **1.96** | **1.0** | **2.91** | **2.62** |

## 5.6 Housekeeping Segmentation

In this section, we claim that our original dataset can be used for housekeeping segmentation tasks. Our original dataset has about 1864 image pairs. Each pair includes a raw poor housekeeping image, a raw good housekeeping image (unaligned), and ground truth. We conduct ablation experiments in Tab.5 and comparison experiments Tab.6. We find that when we use unaligned good housekeeping images, the performance is similar as the performance without good housekeeping images. The difference in Tab.5 is smaller than the difference in Tab.4. It means that, when we use unaligned good housekeeping images, because the unaligned good housekeeping images are diverse, the model degenerates to a segmentation model. This degenerated model seems to only consider the poor housekeeping image, ignoring the diverse good housekeeping image. In other words, we only need to feed the poor housekeeping image and a black box image into the model and the model can still segment the poor housekeeping image, even achieving a good performance of about 86.0% accuracy. In summary, the original dataset can be used as a segmentation dataset for locating and segmenting the causes which result in poor housekeeping.

Table. 5 The segmentation performance with different inputs

| Metric | aACC | mF-score | mPrecision | mRecall | mIoU |
|---|---|---|---|---|---|
| Input without poor images | 70.74 | 65.37 | 64.91 | 66.16 | 50.09 |
| Input without good images | 85.3 | 82.26 | 81.59 | 83.06 | 70.53 |
| Input with all images | 86.0 | 82.91 | 82.53 | 83.32 | 71.46 |
| Difference | **0.7** | **0.65** | **0.94** | **0.26** | **0.93** |

Table. 6 The segmentation performance using SOTA methods on our original dataset.

| Metric | year | Backbone | Lr schd | aACC | mF-score | mPrecision | mRecall | mIoU |
|---|---|---|---|---|---|---|---|---|
| Changer | 2023 | Resnet18 | 40000 | 80.32 | 76.23 | 75.71 | 76.88 | 62.63 |
| Changeformer | 2022 | MIT-B1 | 40000 | 82.6 | 76.79 | 79.73 | 75.07 | 63.78 |
| STANet | 2020 | BAM | 40000 | 73.33 | 61.11 | 65.77 | 60.36 | 47.64 |
| LightCDnet | 2023 | Custom-large | 40000 | 73.2 | 70.33 | 69.79 | 73.54 | 55.04 |
| TTP | 2024 | ViT-SAM-l | 300e | 85.51 | 82.11 | 82.0 | 82.14 | 70.4 |
| Ours | 2024 | Clip, FFM | 40000 | **86.0** | **82.91** | **82.53** | **83.32** | **71.46** |

# 6 Discussion

Housekeeping is a critical task in maintaining safety and efficiency across various workplace environments. Housekeeping aims at creating or maintaining an orderly, clean, tidy, and safe working environment. Effective good housekeeping is essential for preventing injuries and accidents. Conversely, poor housekeeping can introduce hazards that lead to accidents. With the rapid advancements in AI and CV, recent efforts have been made to apply CV models to recognize poor housekeeping practices and prevent accidents. However, explanation models of causes of poor or good housekeeping have not developed. Besides, poor housekeeping segmentation models and datasets have not been well developed. The high-accuracy specifically designed CV models for housekeeping are lacking.

In this study, we introduce the new Housekeeping-CCD dataset, including 700 image pairs (good housekeeping, poor housekeeping and ground truth). Unlike existing construction datasets, Housekeeping-CCD dataset is designed specifically for analyzing housekeeping in construction sites. Different from existing building change detection datasets, Housekeeping-CCD dataset is captured by commodity-grade camera, and our images are captured from a height similar to that of workers, providing a more dynamic and diverse perspective. Images in Housekeeping-CCD are captured with diverse viewpoints in complex scenes. Since we use consumer-grade cameras, with shooting heights ranging from 0.5m to 3m and diverse shooting angles, this can bring in a wide range of practical applications. We also propose a dataset filtering method to refine and clean poor/good housekeeping change detection data.

To improve the accuracy of change detection approaches, we have designed a change detection neural network (HCDN) that integrates CLIP and a proposed feature fusion module, achieving state-of-the-art performance. Compared with other methods, our HCDN provides a new baseline in the change detection field. Indeed, our HCDN can be used in other change detection tasks. To enhance more practical value, we also established a housekeeping segmentation dataset and compare it against state-of-the-art methods to locate and segment objects which cause the poor housekeeping issue.

This study has practical value and applications. (1) Safety monitoring: detects poor housekeeping causes that could lead to accidents, enabling timely interventions. It can help in maintaining a safer work environment and reducing the risk of incidents such as slips, trips, falls, etc. (2) Waste management: detects and segments waste within the workplace. By accurately identifying waste and its locations, our system facilitates more efficient waste removal processes for managers. It can help in keeping the workplace clean and support environmental sustainability efforts. (3) Worker training: uses dataset and network outputs to train workers on best practices for maintaining clean and safe work environments.

Fig 10 illustrates a process flow for poor housekeeping change detection system designed to enhance safety and cleanliness in construction sites. The process begins with the collection of recorded footage from site body-worn cameras or phone cameras, which are strategically used by personnel working on the site. These cameras continuously capture video as workers or robots go about their tasks, ensuring complete coverage of the built environment. Once this video is obtained, the first step is frame grabbing. In this step, specific frames are extracted from the continuous video. After the frames have been successfully grabbed, they are fed into our computer vision model (HCDN). Our model is tasked with the crucial job of change detection, where it identifies images that indicate poor housekeeping. The final step in the process is communication. The results from our HCDN—essentially images highlighting the identified hazards—are sent directly to managers and other relevant personnel through a messaging application, specifically via a Telegram channel. Telegram is chosen for its robust features, including real-time messaging, secure communication, and the ability to easily distribute information to multiple users. By using this platform, our system ensures that site managers receive timely notifications about housekeeping issues, allowing them to take swift action to mitigate risks. In summary, our system, integrating cameras, advanced computer vision, and real-time messaging, creates a useful tool for maintaining high standards of safety and cleanliness in construction sites. Our system not only can reduce the likelihood of human error in identifying hazards but also can enhance the speed and efficiency with which these hazards are addressed, for achieving a safer and productive construction site.

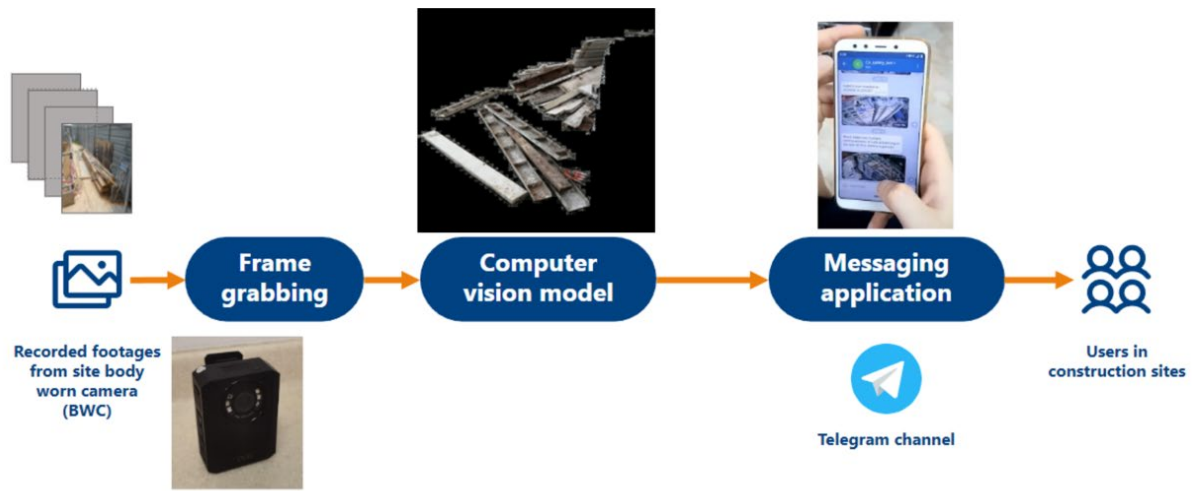

Fig 10. An application system using our change detection method for housekeeping

## 7 Conclusions

Workplace safety is an increasingly critical concern. In construction, poor housekeeping is one of the leading contributors to workplace hazards and accidents. Effective housekeeping is essential to improving WSH, as poor housekeeping can lead to a wide range of dangers, including slips, trips, falls, and falling objects. This paper introduced a change detection dataset, named Housekeeping-CCD, specifically designed for change detection tasks related to construction site housekeeping. Our dataset includes comprehensive annotations to capture various housekeeping conditions, from well-maintained areas to those cluttered with materials, tools, and debris. An improved change detection network named Housekeeping Change Detection Network (HCDN) is proposed, and this incorporates a feature fusion module and a visual large model to achieve superior performance compared to state-of-the-art SOTA methods. Our work also introduces a housekeeping segmentation dataset to detect and segment the specific causes of poor housekeeping. Our approach has significant practical value, not only in improving housekeeping practices but also in its potential applications to other change detection domains beyond construction site housekeeping (e.g., facility management). To promote further development, we share our source code and trained models for global researchers: https://github.com/NUS-DBE/Housekeeping-CD.

In terms of limitations, as highlighted in Fig.8, HCDN can make detection errors, especially when detecting water ponding. The error may be reduced with effective post processing methods, e.g., image filtering. In the future, we will consider the practical considerations for deploying our models at scale. The network training is slow because of the large CLIP model. When deploying our models, we may turn to model compression technologies, including network pruning, model quantization, and knowledge distillation. Network pruning reduces parameters by removing unnecessary connections, convolution kernels, or channels between neurons. Model quantization reduces the model's size and speeds up the calculation by mapping the weight of 32-bit floating-point numbers to 16-bit floating-point numbers, 8-bit integers, etc. Besides, based on the "teacher-student" training method, the knowledge of the large model is "distilled" to the small model so that the small model can perform as close as possible to the large model.

While image-based approaches for construction site safety have made significant strides, they also present other limitations. Firstly, the construction site housekeeping involves both classification and object detection tasks to identify the causes leading to site hazards. As such, expanding new networks that encompassed hybrid classification and object detection can be considered as a part of the future work. In addition, image-based approaches often face challenges such as poor visibility, varying lighting conditions, and the dynamic nature of construction environments. These issues can lead to inaccuracies in detecting and localizing poor housekeeping conditions, particularly in the presence of visual obstructions or adverse weather. Moreover, existing methods may not always provide detailed explanations for their decisions, which is crucial for understanding and addressing the underlying causes of poor housekeeping. It is important to recognize that image-based solutions alone may not be sufficient for comprehensive housekeeping and site safety management. Enhancing the current approach with additional sensor-based technologies, such as LiDAR, and incorporating advanced AI algorithms could further improve the accuracy and applicability of the change detection system. Hybrid systems that integrate multiple data sources are likely to overcome some of the inherent limitations of visual-only methods, offering a more complete view of site conditions and ultimately improving overall safety outcomes.

# 8 Acknowledgement

This research is supported by the National Research Foundation, Singapore under its AI Singapore Programme (AISG Award No:AISG2-100E-2021-084) and the Singapore Housing and Development Board (HDB). We would like to acknowledge that computational work involved in this research work is partially/fully supported by NUS IT's Research Computing group using grant numbers NUSREC-HPC-00001.

# 9 References

[1] 'ILO'. [Online]. Available: https://www.who.int/news/item/17-09-2021-who-ilo-almost-2-million-people-die-from-work-related-causes-each-year

[2] A. S. Sánchez, P. R. Fernández, F. S. Lasheras, F. J. de Cos Juez, and P. G. Nieto, 'Prediction of work-related accidents according to working conditions using support vector machines', *Applied Mathematics and Computation*, vol. 218, no. 7, pp. 3539–3552, 2011.

[3] H. Liu, V. J. L. Gan, J. C. P. Cheng, and S. (Alexander) Zhou, 'Automatic Fine-Grained BIM element classification using Multi-Modal deep learning (MMDL)', *Advanced Engineering Informatics*, vol. 61, p. 102458, Aug. 2024, doi: 10.1016/j.aei.2024.102458.

[4] R. A. Owiti, M. Wandolo, and T. Kinuthia, 'WORK ENVIRONMENT OF HOUSEKEEPING EMPLOYEES ON JOB PERFORMANCE, IN 3-5 STAR HOTELS IN NAIROBI CITY COUNTY, KENYA', *African Journal of Emerging Issues*, vol. 6, no. 11, pp. 118–129, 2024.

[5] 'Workplace Safety and Health Guidelines'. [Online]. Available: https://www.tal.sg/wshc/-/media/tal/wshc/resources/publications/wsh-guidelines/files/wsh-guidelines-on-workplace-housekeeping.ashx

[6] F. Emuze, L. Seboka, and M. Linake, 'Construction work and the housekeeping challenge in Lesotho', Association of Researchers in Construction Management: 32nd Annual ARCOM …, 2016.

[7] E. Aboagye-Nimo and F. Emuze, 'Construction safety through housekeeping: The Hawthorne effect', *Journal of construction project management and innovation*, vol. 7, no. 2, pp. 2027–2038, 2017.

[8] 'Video surveillance system (VSS) for construction sector'. [Online]. Available: https://www.mom.gov.sg/workplace-safety-and-health/safe-measures/sectoral-level/video-surveillance-system-for-construction-sector

[9] The Government of the Hong Kong Special Administrative Region, 'Smart Site Safety System'. [Online]. Available: https://www.devb.gov.hk/filemanager/technicalcirculars/en/upload/1393/1/C-2023-03-01.pdf

[10] Y. G. Lim, J. Wu, Y. M. Goh, J. Tian, and V. Gan, 'Automated classification of "cluttered" construction housekeeping images through supervised and self-supervised feature representation learning', *Automation in Construction*, vol. 156, p. 105095, 2023.

[11] A. Xuehui, Z. Li, L. Zuguang, W. Chengzhi, L. Pengfei, and L. Zhiwei, 'Dataset and benchmark for detecting moving objects in construction sites', *Automation in Construction*, vol. 122, p. 103482, 2021.

[12] R. Duan, H. Deng, M. Tian, Y. Deng, and J. Lin, 'SODA: A large-scale open site object detection dataset for deep learning in construction', *Automation in Construction*, vol. 142, p. 104499, Oct. 2022, doi: 10.1016/j.autcon.2022.104499.

[13] X. Yan, H. Zhang, Y. Wu, C. Lin, and S. Liu, 'Construction Instance Segmentation (CIS) Dataset for Deep Learning-Based Computer Vision', *Automation in Construction*, vol. 156, p. 105083, Dec. 2023, doi: 10.1016/j.autcon.2023.105083.

[14] M. Wang, G. Yao, Y. Yang, Y. Sun, M. Yan, and R. Deng, 'Deep learning-based object detection for visible dust and prevention measures on construction sites', *Developments in the Built Environment*, vol. 16, p. 100245, Dec. 2023, doi: 10.1016/j.dibe.2023.100245.

[15] Xiao Bo and Kang Shih-Chung, 'Development of an Image Data Set of Construction Machines for Deep Learning Object Detection', *Journal of Computing in Civil Engineering*, vol. 35, no. 2, p. 05020005, Mar. 2021, doi: 10.1061/(ASCE)CP.1943-5487.0000945.

[16] B. Xiao, H. Xiao, J. Wang, and Y. Chen, 'Vision-based method for tracking workers by integrating deep learning instance segmentation in off-site construction', *Automation in Construction*, vol. 136, p. 104148, Apr. 2022, doi: 10.1016/j.autcon.2022.104148.

[17] K.-S. Kang, Y.-W. Cho, K.-H. Jin, Y.-B. Kim, and H.-G. Ryu, 'Application of one-stage instance segmentation with weather conditions in surveillance cameras at construction sites', *Automation in Construction*, vol. 133, p. 104034, Jan. 2022, doi: 10.1016/j.autcon.2021.104034.

[18] W. Lu, J. Chen, and F. Xue, 'Using computer vision to recognize composition of construction waste mixtures: A semantic segmentation approach', *Resources, Conservation and Recycling*, vol. 178, p. 106022, Mar. 2022, doi: 10.1016/j.resconrec.2021.106022.

[19] D. Sirimewan, M. Bazli, S. Raman, S. R. Mohandes, A. F. Kineber, and M. Arashpour, 'Deep learning-based models for environmental management: Recognizing construction, renovation, and demolition waste in-the-wild', *Journal of Environmental Management*, vol. 351, p. 119908, Feb. 2024, doi: 10.1016/j.jenvman.2023.119908.

[20] J. Hwang, J. Kim, S. Chi, and J. Seo, 'Development of training image database using web crawling for vision-based site monitoring', *Automation in Construction*, vol. 135, p. 104141, Mar. 2022, doi: 10.1016/j.autcon.2022.104141.

[21] M. M. Woldeamanuel, T. Kim, S. Cho, and H.-K. Kim, 'Estimation of concrete strength using thermography integrated with deep-learning-based image segmentation: Case studies and economic analysis', *Expert Systems with Applications*, vol. 213, p. 119249, Mar. 2023, doi: 10.1016/j.eswa.2022.119249.

[22] M. Kamari and Y. Ham, 'AI-based risk assessment for construction site disaster preparedness through deep learning-based digital twinning', *Automation in Construction*, vol. 134, p. 104091, Feb. 2022, doi: 10.1016/j.autcon.2021.104091.

[23] A. Ji, A. W. Z. Chew, X. Xue, and L. Zhang, 'An encoder-decoder deep learning method for multi-class object segmentation from 3D tunnel point clouds', *Automation in Construction*, vol. 137, p. 104187, May 2022, doi: 10.1016/j.autcon.2022.104187.

[24] A. Pal, J. J. Lin, S.-H. Hsieh, and M. Golparvar-Fard, 'Activity-level construction progress monitoring through semantic segmentation of 3D-informed orthographic images', *Automation in Construction*, vol. 157, p. 105157, Jan. 2024, doi: 10.1016/j.autcon.2023.105157.

[25] R. Huang, Y. Xu, L. Hoegner, and U. Stilla, 'Semantics-aided 3D change detection on construction sites using UAV-based photogrammetric point clouds', *Automation in Construction*, vol. 134, p. 104057, Feb. 2022, doi: 10.1016/j.autcon.2021.104057.

[26] T. Meyer, A. Brunn, and U. Stilla, 'Change detection for indoor construction progress monitoring based on BIM, point clouds and uncertainties', *Automation in Construction*, vol. 141, p. 104442, Sep. 2022, doi: 10.1016/j.autcon.2022.104442.

[27] T. Patel, B. H. Guo, J. D. van der Walt, and Y. Zou, 'Unmanned ground vehicle (UGV) based automated construction progress measurement of road using LSTM', *Engineering, Construction and Architectural Management*, 2024.

[28] T. Czerniawski, J. W. Ma, and Fernanda Leite, 'Automated building change detection with amodal completion of point clouds', *Automation in Construction*, vol. 124, p. 103568, Apr. 2021, doi: 10.1016/j.autcon.2021.103568.

[29] S. Shirowzhan, S. M. E. Sepasgozar, H. Li, J. Trinder, and P. Tang, 'Comparative analysis of machine learning and point-based algorithms for detecting 3D changes in buildings over time using bi-temporal lidar data', *Automation in Construction*, vol. 105, p. 102841, Sep. 2019, doi: 10.1016/j.autcon.2019.102841.

[30] J. Zhang *et al.*, 'AERNet: An Attention-Guided Edge Refinement Network and a Dataset for Remote Sensing Building Change Detection', *IEEE Transactions on Geoscience and Remote Sensing*, vol. 61, pp. 1–16, 2023, doi: 10.1109/TGRS.2023.3300533.

[31] H. Chen and Z. Shi, 'A spatial-temporal attention-based method and a new dataset for remote sensing image change detection', *Remote Sensing*, vol. 12, no. 10, p. 1662, 2020.

[32] S. Ji, S. Wei, and M. Lu, 'Fully Convolutional Networks for Multisource Building Extraction From an Open Aerial and Satellite Imagery Data Set', *IEEE Transactions on Geoscience and Remote Sensing*, vol. 57, no. 1, pp. 574–586, Jan. 2019, doi: 10.1109/TGRS.2018.2858817.

[33] L. Shen *et al.*, 'S2Looking: A satellite side-looking dataset for building change detection', *Remote Sensing*, vol. 13, no. 24, p. 5094, 2021.

[34] S. Fang, K. Li, and Z. Li, 'Changer: Feature Interaction is What You Need for Change Detection', *IEEE Transactions on Geoscience and Remote Sensing*, vol. 61, pp. 1–11, 2023, doi: 10.1109/TGRS.2023.3277496.

[35] Y. Xing, J. Jiang, J. Xiang, E. Yan, Y. Song, and D. Mo, 'Lightcdnet: Lightweight change detection network based on vhr images', *IEEE Geoscience and Remote Sensing Letters*, 2023.

[36] K. Chen *et al.*, 'Time travelling pixels: Bitemporal features integration with foundation model for remote sensing image change detection', *arXiv preprint arXiv:2312.16202*, 2023.

[37] K. Li, X. Cao, and D. Meng, 'A new learning paradigm for foundation model-based remote-sensing change detection', *IEEE Transactions on Geoscience and Remote Sensing*, vol. 62, pp. 1–12, 2024.

[38] A. Radford *et al.*, 'Learning transferable visual models from natural language supervision', presented at the International conference on machine learning, PMLR, 2021, pp. 8748–8763.

[39] A. Vaswani *et al.*, 'Attention is all you need', *Advances in neural information processing systems*, vol. 30, 2017.

[40] W. G. C. Bandara and V. M. Patel, 'A transformer-based siamese network for change detection', presented at the IGARSS 2022-2022 IEEE International Geoscience and Remote Sensing Symposium, IEEE, 2022, pp. 207–210.

[41] K. Sun, 'DMFF: Deep multimodel feature fusion for building occupancy detection', *Building and Environment*, vol. 253, p. 111355, Apr. 2024, doi: 10.1016/j.buildenv.2024.111355.